\documentclass[11pt]{article}
\usepackage[table,xcdraw]{xcolor}
\usepackage{float}
\usepackage{coling2020}
\usepackage{geometry}
\usepackage{times}
\usepackage{url}
\usepackage{hyperref}
\usepackage{latexsym}
\usepackage{microtype}
\usepackage{graphicx}
\usepackage{multirow}

\usepackage{soul}

\colingfinalcopy 

\title{IITK at SemEval-2020 Task 8:  Unimodal and Bimodal Sentiment Analysis of Internet Memes}

\author{Vishal Keswani$^{*}$ \qquad    
  Sakshi Singh\thanks{\quad Authors equally contributed  to this work.}  \qquad  
  Suryansh Agarwal \qquad
  Ashutosh Modi  \\
{Indian Institute of Technology Kanpur (IITK)} \\
  {\tt \{vkeswani,sakshia,asurya\}@iitk.ac.in}  \\
  {\tt ashutoshm@cse.iitk.ac.in}  \\
}
\date{}

\begin{document}
\maketitle
\begin{abstract}
Social media is abundant in visual and textual information presented together or in isolation. Memes are the most popular form, belonging to the former class. In this paper, we present our approaches for the Memotion Analysis problem as posed in SemEval-2020 Task 8. The goal of this task is to classify memes based on their emotional content and sentiment. We leverage techniques from Natural Language Processing (NLP) and Computer Vision (CV) towards the sentiment classification of internet memes (Subtask A). We consider Bimodal (text and image) as well as Unimodal (text-only) techniques in our study ranging from the Naïve Bayes classifier to Transformer-based approaches. Our results show that a text-only approach, a simple Feed Forward Neural Network (FFNN) with Word2vec embeddings as input, performs superior to all the others. We stand first in the Sentiment analysis task with a relative improvement of 63\% over the baseline macro-F1 score. Our work is relevant to any task concerned with the combination of different modalities.
\end{abstract}

\section{Introduction}
\blfootnote{
    %
    %
    %
    %
    %
    
     \hspace{-0.65cm}  
     This work is licensed under a Creative Commons 
     Attribution 4.0 International License.
     License details:
     \url{http://creativecommons.org/licenses/by/4.0/}.
}
An internet meme conveys an idea or phenomenon that is replicated, transformed and spread through the internet. Memes are often grounded in personal experiences and are a means of showing appeal, resentment, fandom, along with socio-cultural expression. Nowadays, the popularity of the internet memes culture is on a high. This creates an opportunity to draw meaningful insights from memes to understand the opinions of communal sections of society. The inherent sentiment in memes has political, social and psychological relevance. The sarcasm and humor content of memes is an indicator of many cognitive aspects of users. Social media is full of hate speech against communities, organizations as well as governments. Analysis of meme content would help to combat such societal problems.  

The abundance and easy availability of multimodal data have opened many research avenues in the fields of NLP and CV. Researchers have been working on analyzing the personality and behavioral traits of humans based on their social media activities, mainly posts they share on Facebook, Twitter, etc. \cite{golbeck2011twitter}. In this regard, we attempt to solve the sentiment classification problem under SemEval-2020 Task 8: "Memotion Analysis" \cite{chhavi2020memotion}. Memotion analysis stands for analysis of emotional content of memes. We dissociate the visual and textual components of memes and combine information from both modalities to perform sentiment-based classification. The literature is rich in sentiment classification for tweets \cite{sailunaz2019emotion} and other text-only tasks \cite{devlin2018bert}. The multimodal approaches are relatively recent and yet under exploration \cite{morency2017multimodal,cai2015convolutional,kiela2019supervised}. 
 
 We first describe the problem formally in section \ref{sec:background}, followed by a brief literature review of the work already done in this domain. In section \ref{sec:methods}, we describe the methods proposed by us. Section \ref{sec:expsetup} contains the description of the dataset provided by the organizers, along with the challenges accompanying the data. It further takes a deeper dive into the method yielding the best results. Section \ref{sec:results} summarizes the results with a brief error analysis. Towards the end, section \ref{sec:conclusion} concludes the paper along with future directions. The implementation for our system is made available via Github\footnote{\url{https://github.com/vkeswani/IITK_Memotion_Analysis}}

\begin{figure}[]
\centering
\includegraphics[trim= 0 145 0 0, clip, scale=0.55]{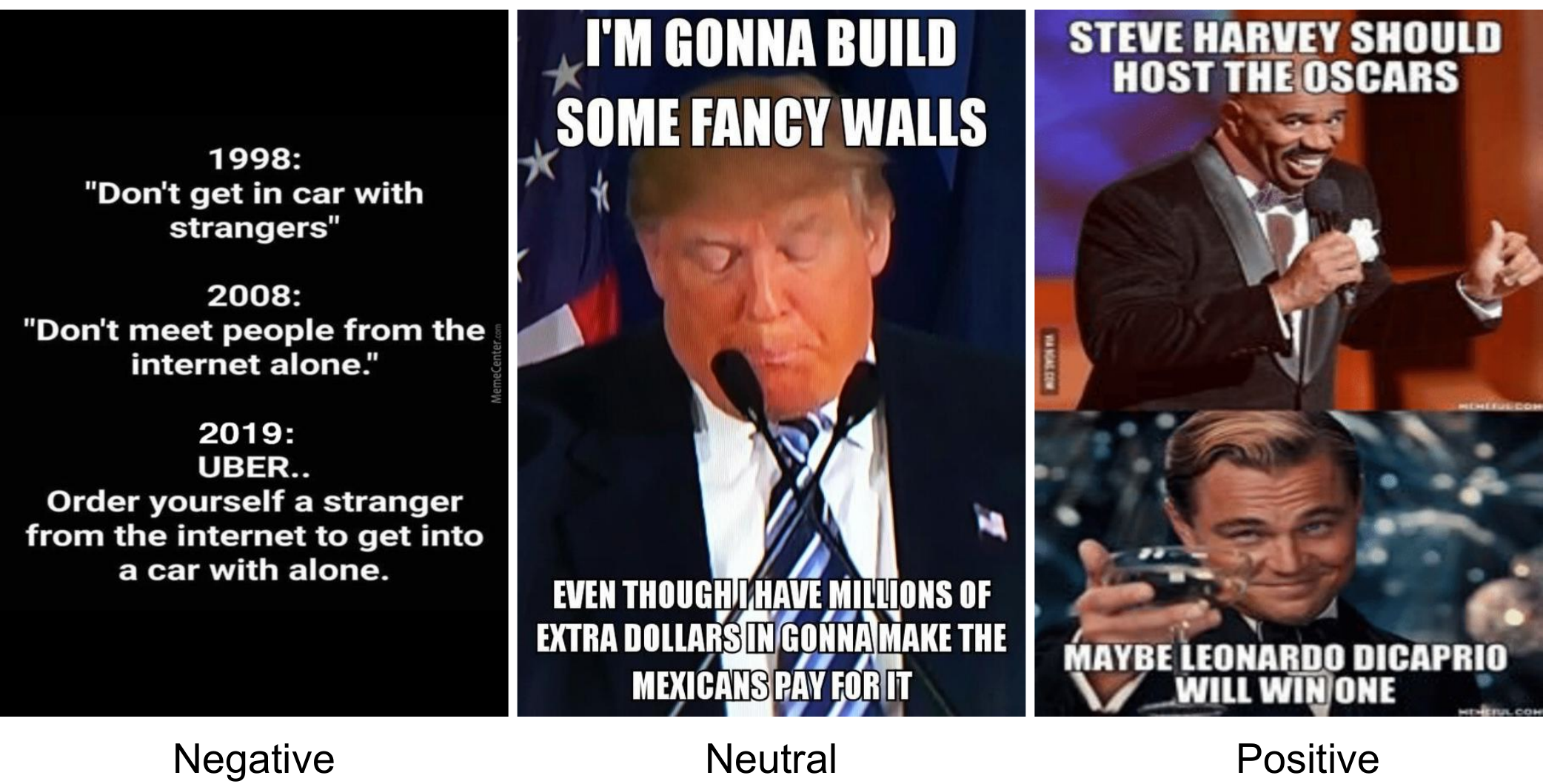}
\label{fig:example}
\caption{Pipeline for FFNN with Word2Vec}
\end{figure}
 
 \section{Background}
 \label{sec:background}
Subtask A of Memotion Analysis is about  Sentiment Analysis of memes. It is concerned with the classification of an Internet meme as a positive, negative, or neutral sentiment. Figure \ref{fig:example} illustrates the examples of these three classes.

A major portion of the research is devoted to separate handling of text and image modalities. Among the text-only methods, the Transformer \cite{vaswani:17}  is an encoder-decoder architecture that uses only the attention mechanism instead of RNN. It has wide applications in different NLP tasks. BERT \cite{devlin2018bert} is the state-of-the-art transformer model for text only classification. It introduced the concept of bidirectionality to the attention mechanism to allow better use of contextual information. Reformer \cite{kitaev2020reformer} is the most recent transformer, but more memory efficient and faster, hence works better for long sequences. 

For image-based tasks, ImageNet \cite{oquab2014learning} is a visual dataset with hand annotated images for visual object recognition task. ResNet-152 \cite{he2016deep} is a deep learning model trained using the Imagenet dataset and used for object classification.

The joint handling of image and text modalities has gained attention relatively recently. \newcite{qiantext} proposed a Text-Image Sentiment Analysis model that linearly combines image and text features. MMBT \cite{kiela2019supervised} or the Multimodal Bitransformer fuses information from text and image encoders (BERT and ResNet) by mapping the image embeddings into the text space. 

\section{Methods}
\label{sec:methods}
A wide variety of methods, ranging from a simple linear classifier to transformers, were employed. We broadly classify the set of techniques into bi-modal and uni-modal.

\subsection{Bi-modal methods} These approaches consider both text and image modalities for classification. In general, both modalities are first treated separately, and high-level features are derived. These features are then combined using an additional classifier to make the final prediction. We use two main approaches:

\subsubsection{Text-only FFNN and Image-only CNN:} As proposed in \newcite{qiantext}, this method uses FFNN for text analysis (one-hot encoding for vectorization) and CNN for images analysis (HSV values for vectorization). For both the analysis, we get a probability distribution over the classes (positive, negative and neutral) as output. We concatenate the predicted probability distributions from the above two models and feed them as features to an additional classifier (SVM in this case) to arrive at a final prediction.

\subsubsection{Multimodal Bitransformer (MMBT):} MMBT \cite{kiela2019supervised} is a recent advancement in the fusion of unimodal encoders. They are individually pre-trained in a supervised fashion. It combines ResNet-152 and BERT by mapping the image embeddings into the text space, followed by a classification layer. It is a flexible architecture and works even if one modality is missing and captures text dominance. It can also handle arbitrary lengths of inputs and an arbitrary number of modalities. We fine-tuned MMBT for our dataset. 

\subsection{Uni-modal methods} We experimented with three text-only approaches for meme classification. The emphasis on separate text-only analysis is justified in subsection \ref{subsec:data}.

\subsubsection{Na\"ive Bayes}
Na\"ive Bayes is a popular classical machine learning classifiers \cite{rish2001empirical}. The main assumption behind the model is that given the class labels. All features are conditionally independent of each other, hence the name Na\"ive Bayes. It is highly scalable, that is, takes less training time. It also works well on small datasets, making it a good baseline for our analysis. We used the default implementation of Na\"ive Bayes classifier provided by the TextBlob library\footnote{https://textblob.readthedocs.io/en/dev/} \cite{loria2014textblob}.

\subsubsection{Text-only FFNN}
\label{subsec:ann}
We use Word2vec embeddings \cite{mikolov2013efficient} for capturing semantic and syntactic properties of words. It is a dense low-dimensional representation of a word. We use the pre-trained embeddings. Word2Vec represents each word as a vector (1x300 in our case). A caption is represented by an average of word embeddings of each of the words. Consequently, the input to FFNN is an $n \times 300$ matrix, where $n$ is the number of captions.

\subsubsection{BERT} Bidirectional Encoder Representations from Transformers (BERT) \cite{devlin2018bert} is the state-of-the-art language model, that has been found to be useful for numerous NLP tasks. It is deeply bidirectional (takes contextual information from both sides of the token) and learns a representation of text via self-supervised learning. BERT models pre-trained on large text corpora are available, and these can be fine-tuned for a specific NLP task. We fine-tuned BERT Base Uncased configuration\footnote{https://github.com/google-research/bert}, which has 12 layers (transformer blocks), 12 attention heads and 110 million parameters.

\section{Experimental Setup}
\label{sec:expsetup}
In this section, we quantitatively describe the dataset provided by the organizers and challenges accompanying it. We then mention the preprocessing steps briefly. Finally, we discuss the architecture and parameters of the FFNN with Word2vec approach (Section \ref{subsec:ann}) in detail. This method made it to the final submission as it performed better than all the other approaches.  

\subsection{Data description}
\label{subsec:data}
As a part of the task, we are provided with 7K human-annotated Internet memes labelled with various semantic dimensions (positive, negative or neutral for subtask A). The dataset also contains the extracted captions/texts from the memes using the Google OCR system and then manually corrected by crowdsourcing services. Hence, we have two modalities, image and text.  

The dataset (Table \ref{tab:Datasettable}) comes with a lot of inherent challenges. Firstly, the sentiment or emotion perceived depends on the social or professional background of the perceiver. Hence, classification is highly subjective. Also, the presence of sarcasm in memes makes sentiment classification a difficult task since positive appearing features are grounded in negative sentiment by the application of sarcasm. The large variance in caption lengths is another issue. The text sequence lengths vary from 1 (or even 0) to 100+.

Also, the text dominance in the memes provided is evident from glimpsing the data as the same image-template is repeated across different classes due to popularity of the image and bear very low correlation with the sentiment class. Hence, a good approach takes text as the dominant modality, and text-only approaches work well. 

\begin{table}[H]
\centering
\renewcommand{\arraystretch}{1.1}%
\begin{tabular}{|c|c|c|c|}
\hline
\rowcolor[HTML]{D4D0D0} 
\textbf{Dataset} & \textbf{Class} & \textbf{Points} & \textbf{Percentage} \\ \hline
               & Positive       & 4160   & 59.5\%     \\ \cline{2-4}  
                           & Neutral & 2201 & 31.5\% \\  \cline{2-4}
               & Negative       & 631      & 9.0\%       \\ \cline{2-4}
\multirow{-4}{*}{Training} & Total   & 6992    & 100\% \\ \hline
Test           & Total          & 1788        & 25.6\%    \\ \hline
\end{tabular}
\caption{Class distribution for subtask A}
\label{tab:Datasettable}
\end{table}

\subsection{Data preprocessing}
Text preprocessing steps included removal of punctuation, stop words and special characters, followed by lower-casing, lemmatization and tokenization. We used nltk library\footnote{https://pythonspot.com/category/nltk/} \cite{loper2002nltk} for the same. The tokens were then converted to vectors using Word2vec embeddings. Finally, the average of all the word vectors is taken to create caption embeddings (as mentioned in section \ref{subsec:ann}).

\begin{figure}[H]
\centering
\includegraphics[trim= 0 143 0 0, clip, scale=0.75]{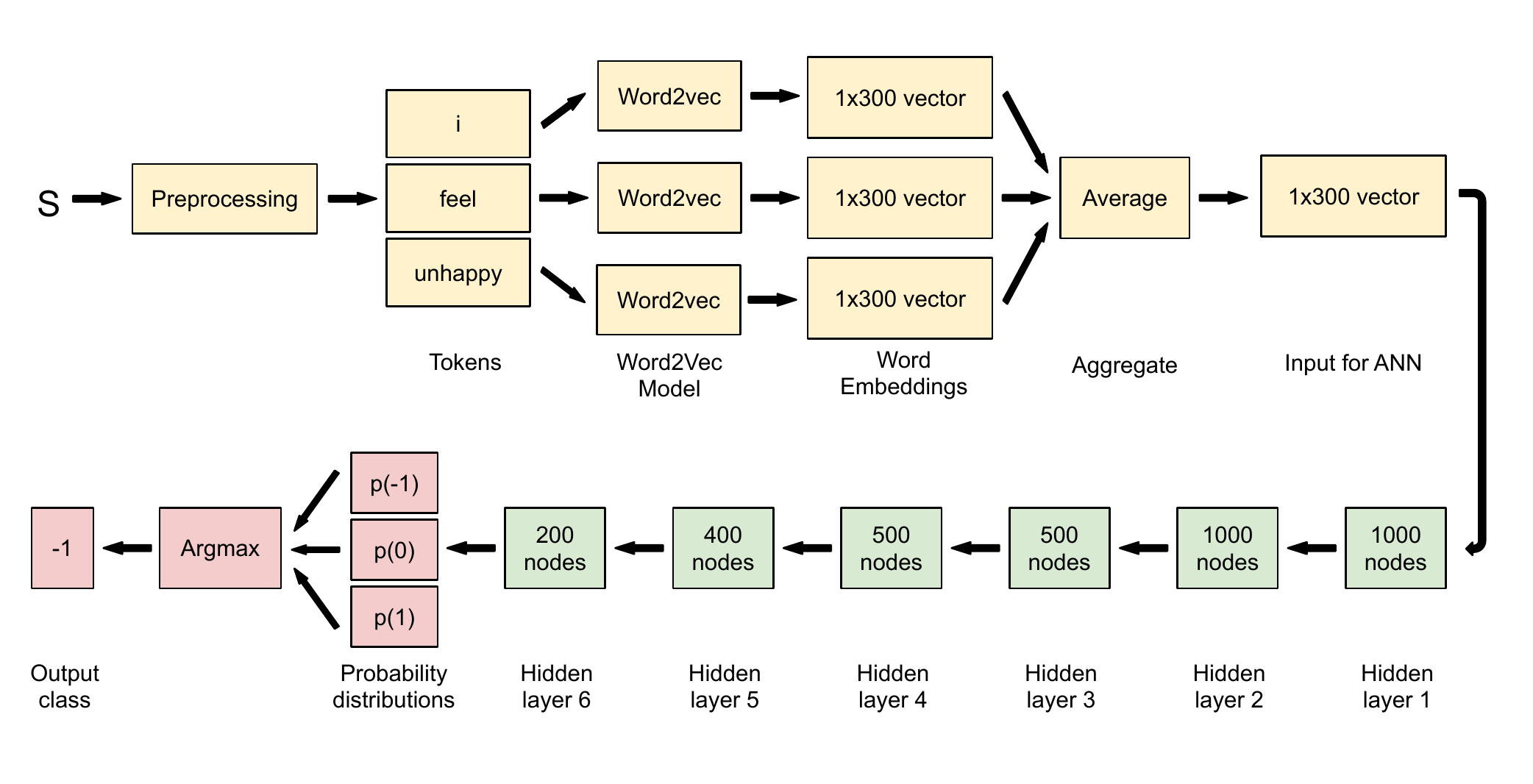}
\caption{Pipeline for FFNN with Word2Vec. S="I am feeling unhappy."}
\label{fig:flowchart}
\end{figure}
\subsection{Model and parameters}
We elaborate more upon the precise architecture of the text-only FFNN approach with Word2vec embeddings. We employ a Feed-Forward Neural Network, with 6 hidden layers, softmax cross-entropy, Adam optimizer \cite{kingma2014adam} and ReLu activation function. The number of nodes in each hidden layer is depicted in Figure \ref{fig:flowchart}. Weights for each hidden layer are initialized with the standard normal distribution. The batch size is 50, and the number of epochs is 10. 

FFNN is susceptible to randomness due to various reasons like random initialization of weight matrices, randomness in optimization, etc. Hence, it gives different results on multiple runs with the same parameters/hyperparameters. For a larger dataset (unlike ours), the model may produce more stable results. In the following section, we present the best score on the test set (Table \ref{tab:results}) along with the mean and the variance of the scores on the validation set for 50 runs (Table \ref{tab:results2}).

\begin{table}[H]
\centering
\begin{tabular}{|c|c|c|c|}
\hline
Mean        & Variance    & Max         & No. of runs \\ \hline
0.34     & 2E-4      & 0.36  & 50   \\ \hline
\end{tabular}
\caption{Macro-F1 for text-only FFNN (Word2vec) on validation set (80:20 split)}
\label{tab:results2}
\end{table}

\section{Results}
\label{sec:results}
The official evaluation metric for Memotion Analysis is the Macro-F1. We present the Macro-F1 scores of our five main approaches for subtask A in Table \ref{tab:results}. With the best Macro-F1 of 0.3546581568, we improve the baseline (0.2176489217) by 63\% and are ranked first in subtask A.

The dominance of the majority class over the others played a crucial role in sub-par performance for the other classes. For the `negative' class, there are not enough data-points for the system to train. To some extent, this issue was resolved using simple upsampling. The transformer-based approaches overfit heavily to the majority class. The repetition of meme templates is another issue for the bimodal approaches. Sarcasm also introduced ambiguity in the sentiment classification task. Neutral class dominated relatively due to the absence of polar words in some sarcastic texts.  

Our results may be surprising as the state-of-the-art models, BERT and MMBT, are expected to do better. Some of the underlying reason for such behavior can be fewer data points, sarcasm, noise, and large variance in caption lengths. Moreover, BERT has been pre-trained on Wikipedia and Book Corpus, a dataset containing +10,000 books of different genres. These contained well-defined sentences, but our dataset is noisy, lacks punctuation marks and comprises of sarcasm. Simpler approaches did a better job since they involve no pre-training on any other (large) corpus.

\begin{table}[H]
\centering
\renewcommand{\arraystretch}{1.1}%

\begin{tabular}{|c|c|c|}
\hline
\rowcolor[HTML]{D0D0D0} 
\textbf{Modality}           & \textbf{Model}                          & \textbf{Macro-F1}            \\ \hline
                            & FFNN + CNN  & 0.29 \\  \cline{2-3}
\multirow{-2}{*}{Text-Image} & MMBT & 0.30 \\ \hline
                            & Naive Bayes                             & 0.32            \\  \cline{2-3}
                            & FFNN (Word2Vec)  & \textbf{0.35} \\ \cline{2-3}
\multirow{-3}{*}{Text-Only} & BERT & 0.33  \\ \hline
\multicolumn{2}{|c|}{Baseline} & \multicolumn{1}{c|}{0.22} \\ \hline
\end{tabular}
\caption{Results for subtask A}
\label{tab:results}
\end{table}

\section{Conclusion}
\label{sec:conclusion}
We attempt to perform a complex task of classifying memes constrained by the data size and quality. While the results were sound for the populous classes, it could only be par after re-sampling for the skewed classes. The best results were obtained for FFNN with Word2vec embeddings (Table \ref{tab:results}). Vanilla ANN-based approaches are highly competitive as compared to transformers and even outperformed them. A better research problem would be to define some rules to obtain meme data and then perform the above task so domain knowledge could be used to improve performance. 

In future, a study could be designed to observe the diffusion of memes among different communities by analysing which meme is most liked or hated by a particular community (e.g. a Facebook group). Finding the political inclination of memes is also a possible path. Memes have become an increasingly popular mode of expressing opinions by party supporters, party critics and the affected people. They may be considered as a means of propaganda. This makes the problem of detecting the inclination of memes in a political context an important research exercise. 

\bibliographystyle{coling}
\bibliography{semeval2020}

\end{document}